\documentclass[conference]{IEEEtran}
\IEEEoverridecommandlockouts

\usepackage{cite}
\usepackage{amsmath,amssymb,amsfonts}
\usepackage{algorithmic}
\usepackage{graphicx}
\usepackage{multirow}
\usepackage{textcomp}
\usepackage{url}
\usepackage{xcolor}
\def\BibTeX{{\rm B\kern-.05em{\sc i\kern-.025em b}\kern-.08em
    T\kern-.1667em\lower.7ex\hbox{E}\kern-.125emX}}
\begin{document}

\title{Exploring Large Language Models for Knowledge Graph Completion
}

\author{\IEEEauthorblockN{Liang Yao}
\IEEEauthorblockA{
\textit{Sun Yat-sen University}\\
Shenzhen, China\\
yaoliang3@mail.sysu.edu.cn
}
\and
\IEEEauthorblockN{Jiazhen Peng}
\IEEEauthorblockA{
\textit{Tencent Inc.}\\
Shenzhen, China\\
brucejzpeng@tencent.com }

\and
\IEEEauthorblockN{Chengsheng Mao, Yuan Luo}
\IEEEauthorblockA{
\textit{Northwestern University}\\
Chicago, USA \\
 \{chengsheng.mao, yuan.luo\}@northwestern.edu
}
}

\maketitle

\begin{abstract}
Knowledge graphs play a vital role in numerous artificial intelligence tasks, yet they frequently face the issue of incompleteness. In this study, we explore utilizing Large Language Models (LLM) for knowledge graph completion. We consider triples in knowledge graphs as text sequences and introduce an innovative framework called Knowledge Graph LLM (KG-LLM) to model these triples. Our technique employs entity and relation descriptions of a triple as prompts and utilizes the response for predictions. Experiments on various benchmark knowledge graphs demonstrate that our method attains state-of-the-art performance in tasks such as triple classification and relation prediction. We also find that fine-tuning relatively smaller models (e.g., LLaMA-7B, ChatGLM-6B) outperforms recent ChatGPT and GPT-4.
\end{abstract}

\begin{IEEEkeywords}
Large language model, Knowledge graph completion
\end{IEEEkeywords}

\section{Introduction}
Large knowledge graphs (KG) like FreeBase~\cite{bollacker2008freebase}, YAGO~\cite{suchanek2007yago}, and WordNet~\cite{miller1995wordnet} serve as a powerful foundation for numerous critical AI tasks, including semantic search, recommendation~\cite{zhang2016collaborative}, and question answering~\cite{cui2017kbqa}. A KG is generally a multi-relational graph with entities as nodes and relations as edges. Each edge is depicted as a triplet (\textit{head entity}, relation, \textit{tail entity}) (abbreviated as ($h, r, t$)), signifying the relationship between two entities, for instance, (\textit{Steve Jobs}, founded, \textit{Apple Inc.}). Despite their effectiveness, knowledge graphs remain incomplete. This issue leads to the challenge of \textit{knowledge graph completion}, which aims to evaluate the plausibility of triples that are not present in a knowledge graph.

A significant amount of research has been dedicated to knowledge graph completion. One prevalent method is knowledge graph embedding~\cite{wang2017knowledge}. However, most knowledge graph embedding models solely rely on structural information from observed triple facts, leading to issues arising from the sparsity of knowledge graphs. A number of studies integrate textual information to enhance knowledge representation~\cite{socher2013reasoning,xie2016representation,xiao2017ssp,wang2016text,xu2017knowledge,an2018accurate}. Our previous work KG-BERT~\cite{yao2019kg} firstly employs the pre-trained language model BERT~\cite{devlin2019bert} to encode prior knowledge and contextual information. The KG-BERT model was extended by several recent studies~\cite{wang2021structure,wang2022simkgc,lovelace2022framework,youn-tagkopoulos-2023-kglm} on both efficiency and performance, but the models used in these works are relatively small. 

Recently, large language models~\cite{zhao2023survey} like ChatGPT and GPT-4~\cite{openai2023gpt4} have gained significant attention. Researchers find that scaling pre-trained language models often leads to an improved model capacity on downstream tasks. These large-sized models show different behaviors from smaller models like BERT and display surprising abilities in solving a series of complex tasks.

In this study, we propose a novel method for knowledge graph completion using large language models. Specifically, we treat entities, relations, and triples as textual sequences and model knowledge graph completion as a sequence-to-sequence problem. We perform instruction tuning with open LLMs (LLaMA~\cite{touvron2023llama} and ChatGLM~\cite{du2022glm}) on these sequences for predicting the plausibility of a triple or a candidate entity/relation. The method achieves stronger performance in several KG completion tasks. Our source code is available at:~\url{https://github.com/yao8839836/kg-llm}. Our contributions are summarized as follows:


\begin{itemize}
    \item We propose a new language modeling method for knowledge graph completion. To the best of our knowledge, this is the first study to systematically investigate large language models for KG completion tasks.
    \item Results on several benchmarks show that our method achieves state-of-the-art results in triple classification and relation prediction. We also find that fine-tuning relatively smaller models (e.g., LLaMA-7B, ChatGLM-6B) can outperform recent ChatGPT and GPT-4.
\end{itemize} 

\section{Related Work}
\subsection{Knowledge Graph Completion}
Comprehensive reviews of knowledge graph completion techniques have been carried out by~\cite{wang2017knowledge} and~\cite{ji2021survey}. These techniques can be grouped into two categories based on their scoring functions for triple $(h,r,t)$:  translational distance models like TransE~\cite{bordes2013translating} and semantic matching models like DistMult~\cite{yang2015embedding}. Convolutional neural networks have also demonstrated  promising results in knowledge graph completion~\cite{dettmers2018convolutional,SWJ318,nathani2019learning}.

The methods mentioned above perform knowledge graph completion using only the structural information found in triples. However, incorporating various types of external information, such as entity types, logical rules, and textual descriptions, can enhance performance~\cite{wang2017knowledge,ji2021survey}. For textual descriptions, \cite{socher2013reasoning} initially represented entities by averaging the word embeddings in their names, with the embeddings learned from an external corpus.  \cite{wang2014knowledgeb} suggested embedding entities and words in the same vector space by aligning Wikipedia anchors with entity names. \cite{xie2016representation} employed convolutional neural networks (CNN) to encode word sequences in entity descriptions. There are also a number of studies in this line of works~\cite{xiao2017ssp,wang2016text,xu2017knowledge,an2018accurate}. \cite{yao2019kg} proposed KG-BERT which improves the above methods with pre-trained language models (PLMs). Recently, \cite{wang2021structure,wang2022simkgc,lovelace2022framework} extended cross-encoder in KG-BERT to bi-encoder, which enhances the performance and inference efficiency. Similar to this work, KGT5~\cite{saxena2022sequence} and KG-S2S~\cite{chen2022knowledge} treat KG completion as sequence-to-sequence tasks. However, the pre-trained language models used in these studies are relatively small.

Compared with these methods, our method utilizes more powerful large language models with emergent abilities not present in small models such as in-context learning,  instruction following, and step-by-step reasoning. These abilities are helpful for KG completion tasks.

\subsection{LLMs with KG Completion}
Recently, \cite{zhao2023survey} presents a comprehensive survey of LLMs that describes knowledge completion as a basic evaluation task of LLMs. Three closely related studies~\cite{xie2023lambdakg,zhu2023llms,wei2023kicgpt} evaluate ChatGPT and GPT-4 on a link prediction task in KG. Our study is inspired by these works, but we further provide more comprehensive results for KG completion and perform instruction tuning on three tasks. A recent work~\cite{liu2024finetuning} extended our current work by incorporating structural information from knowledge graph embeddings (e.g., TransE) during instruction fine-tuning. However, they only focused on the link prediction task, while we also explored the tasks of triple classification and relation classification. 

\begin{figure*}[t]
  \centering
  \includegraphics[width = 0.95 \textwidth]{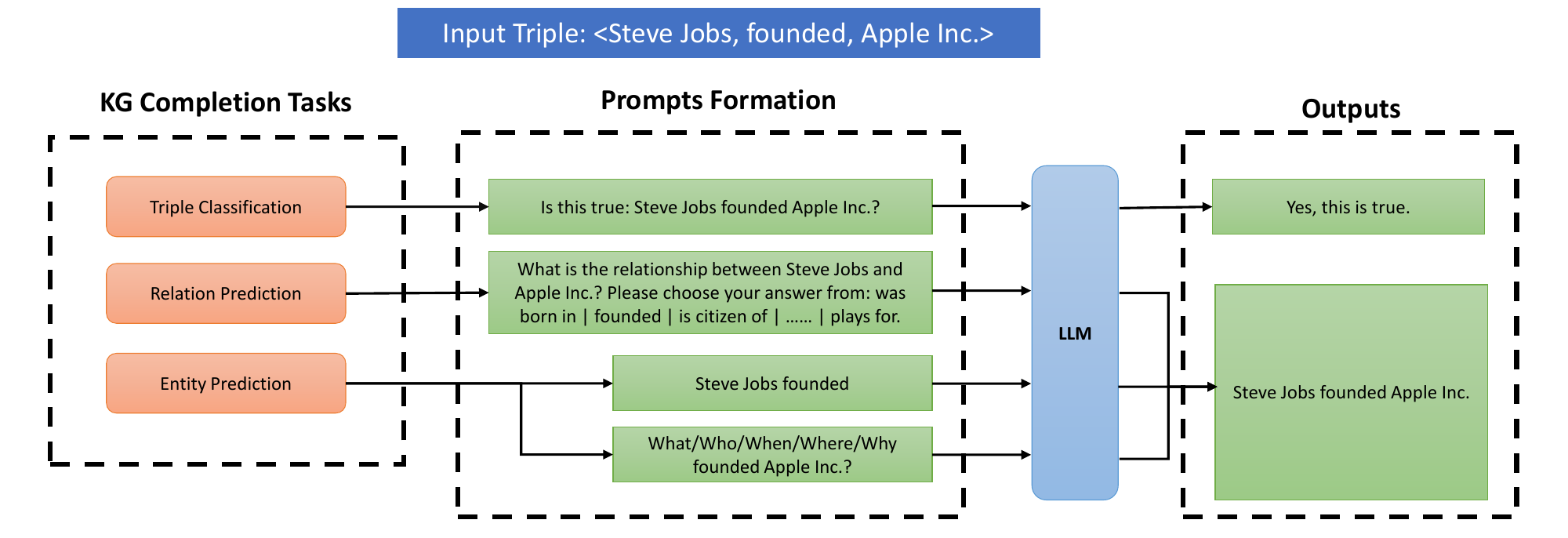}
  \caption{Illustrations of Large Langauge Models (LLMs) for Knowledge Graph (KG) Completion. }
  \label{fig:KG-LLM}
\end{figure*}

\section{Method}




\subsection{Knowledge Graph Completion Tasks}
In this chapter, we describe the three tasks in knowledge graph completion: triple classification, relation prediction, and entity (link) prediction, and how to transform them into simple prompt questions for LLM to complete the tasks. The entire process is depicted in Figure~\ref{fig:KG-LLM}.

\paragraph{Triple Classification.}
Given a triple ($h, r, t$), the task is to classify it as correct or incorrect. For example, given the triple $<$\textit{Steve Jobs}, founded, \textit{Apple Inc.}$>$, the task is to classify it as correct. The prompt formation would be ``Is this true: Steve Jobs founded Apple Inc.?". And the ideal output of LLM would be ``Yes, this is true."

\paragraph{Relation Prediction.}
Given a head entity and a tail entity, the task is to predict the relationship between them. For example, given the head entity ``Steve Jobs" and the tail entity ``Apple Inc.", the task is to predict that their relationship is ``founded".  The prompts formation would be ``What is the relationship between Steve Jobs and Apple Inc.? Please choose your answer from: was born in$|$founded$|$is citizen of$|$……$|$plays for."  And the desired response would be ``Steve Jobs founded Apple Inc."

\paragraph{Entity (link) Prediction.}
Given a head entity and a relationship, the task is to predict the tail entity related to the head entity. Given a tail entity and a relationship, the task is to predict the head entity. For example, given the head entity ``Steve Jobs" and the relationship ``founded", the task is to predict the tail entity ``Apple Inc.". The prompts formation would be ``Steve Jobs founded"  for asking the tail entity and ``What/Who/When/Where/Why founded Apple Inc.?" for asking the head entity. The ideal response would be ``Steve Jobs founded Apple Inc."

\subsection{Instruction Tuning LLM with KG (KG-LLM)}


In order to align LLMs with KG triples, we introduce KG-LLM, which instruction tunes the pre-trained LLM to process KG data using the specific factual question-answering prompt paradigm. Specifically, we fine-tune two open LLMs: ChatGLM-6B~\cite{du2022glm} with P-tuning v2~\cite{liu2021p} and LLaMA (version 1 and 2)~\cite{touvron2023llama} with LoRA~\cite{hu2021lora} using prompts and responses of training triples in a KG. We name our fine-tuned models KG-ChatGLM-6B and KG-LLaMA (7B and 13B). We also incorporate structural information into training and test instructions. Specifically, for the entity prediction task, we sample $K = 5$ neighboring entities (excluding the target entity) for the given entity and tell the model as: ``Giving the neighbors of Steve Jobs: Steve Wozniak$|$USA$|$Bill Gates$|$male$|$California. Complete the fact: Steve Jobs founded".

    \begin{table}[h]
    \caption{Summary statistics of datasets.}
    \centering
    \footnotesize
    \renewcommand{\arraystretch}{1.2}
    \begin{tabular}{c|ccccccc}
    \hline
    \bf{Dataset}& \bf{\# Ent}	& \bf{\# Rel}& \bf{\# Train}& \bf{\# Dev} & \bf{\# Test} \\
    \hline
    WN11 & 38,696 & 11 & 112,581 & 2,609  & 10,544\\ 
    FB13 & 75,043  & 13 & 316,232 & 5,908 & 23,733 \\
    WN18RR& 40,943 & 11 & 86,835 & 3,034 & 3,134\\
    YAGO3-10& 123,182 & 37 & 1,079,040 & 5,000 & 5,000\\
    \hline
    \end{tabular}
    \label{tab:statistics}
    \end{table}

\section{Experiments}

    \begin{table}[t]
      \caption{Triple classification accuracy (in percentage) for different methods. The baseline results with citations are obtained from corresponding papers.}
    \footnotesize
    \centering
    \renewcommand{\arraystretch}{1.1}
    \begin{tabular}{l|cc|c}
    \hline
    \bf{Method}& WN11& FB13 & Avg.\\
    \hline
    NTN~\cite{socher2013reasoning}& 86.2 & 90.0 & 88.1\\
    TransE~\cite{wang2014knowledge} & 75.9 & 81.5 & 78.7\\   
    TransH~\cite{wang2014knowledge} & 78.8 & 83.3 & 81.1\\
    TransR~\cite{lin2015learning} & 85.9 & 82.5  & 84.2\\
    TransD~\cite{ji2015knowledge} & 86.4  & 89.1 & 87.8\\ 
    TEKE~\cite{wang2016text} & 86.1 & 84.2  & 85.2\\ 
    TransG~\cite{xiao2016transg} & 87.4  & 87.3 & 87.4  \\
    TranSparse-S~\cite{ji2016knowledge} & 86.4& 88.2 & 87.3\\
    DistMult~\cite{zhang2018knowledge} & 87.1& 86.2  & 86.7\\
    DistMult-HRS~\cite{zhang2018knowledge} & 88.9& 89.0 & 89.0\\
    AATE~\cite{an2018accurate} & 88.0  & 87.2  & 87.6\\    
    ConvKB~\cite{SWJ318} & 87.6  & 88.8  &  88.2\\
    DOLORES~\cite{wang2018dolores} & 87.5 & 89.3 & 88.4 \\
    DKRL (BERT) & 87.3& 79.8 & 83.6\\    
    KG-BERT(a)~\cite{yao2019kg} & 93.5 & 90.4 & 91.9\\
    KGT5 & 72.8 & 66.3 & 69.6\\
    LLaMA-7B & 21.1 & 9.1 & 15.1\\
    LLaMA-13B & 28.1& 17.6 & 22.9\\
    KG-LLaMA-7B & 95.5 & 89.2 & 92.4\\
    KG-LLaMA-13B & 95.6 & 90.2 & 92.9\\
    KG-LLaMA2-13B & \textbf{96.6} & \textbf{90.7} & \textbf{93.7}\\
    \hline
    \end{tabular}
    \label{tab:triple_results}
    \end{table}
    
    \begin{table}[h]
      \caption{Triple classification accuracy on 100 test instances of FB13 for different LLMs.}
    \footnotesize
    \centering
    \renewcommand{\arraystretch}{1.1}
    \begin{tabular}{l|c}
    \hline
    \bf{Method}& FB13-100 \\
    \hline
    ChatGPT &  0.90  \\
    GPT-4 &  \bf{0.94}  \\    
    LLaMA-7B&  0.14 \\
    LLaMA-13B&  0.16 \\
    KG-LLaMA-7B & 0.93   \\
    KG-LLaMA-13B &  \bf{0.94}  \\
    \hline
    \end{tabular}
  \label{tab:triple_results_100}
   \end{table}

    \begin{table}[t]
      \caption{Entity (link) prediction Hits@1 for different methods. }
    \footnotesize
    \centering
    \renewcommand{\arraystretch}{1.0}
    \begin{tabular}{l|ccc}
    \hline
    \bf{Method}& WN18RR & YAGO3-10 & YAGO3-10-100\\
    \hline
    KG-BERT(a) & 0.1102 &  -- & --\\
    StAR & 0.2430 & -- & --\\
    KGT5 & 0.1011& 0.0484 & 0.12\\
    KGLM & 0.3050 & -- & --\\
    ChatGPT& -- & -- & 0.22\\
    GPT-4& -- & -- & \textbf{0.24}\\
    KG-ChatGLM-6B& 0.1613 & 0.0455 & 0.11\\
    LLaMA-7B  & 0.0849 & 0.0254 & 0.03\\
    LLaMA-13B  & 0.0991 &  0.0276 & 0.01\\
    KG-LLaMA-7B & 0.2415 &  0.0782 & 0.16\\ 
    KG-LLaMA-13B & 0.2559 & 0.0872 & 0.13\\
    KG-LLaMA2-13B & 0.2682 & 0.0949 & 0.16\\
    KG-LLaMA2-13B + 5 Neighbors  & \textbf{0.3151} & \textbf{0.1330}  & 0.22 \\
    \hline
    \end{tabular}
    \label{tab:link_results}
    \end{table}

    \begin{table}[t]
      \caption{Relation prediction Hits@1 scores.}
    \footnotesize
    \centering
    \renewcommand{\arraystretch}{1.1}
    \begin{tabular}{l|c|c}
    \hline
    \bf{Method}& YAGO3-10 & YAGO3-10-100\\
    \hline
    KG-BERT(b) & 0.6816 & --\\
    KGT5 & 0.5714 & 0.60\\
    ChatGPT & -- & 0.39\\
    GPT-4 & -- & 0.56\\
    ChatGLM-6B &  0.0658 & 0.07\\
    KG-ChatGLM-6B& 0.5662 & 0.58\\
    LLaMA-7B &  0.0348 & 0.13\\
    LLaMA-13B &  0.0040 & 0.01\\
    KG-LLaMA-7B &  \textbf{0.7028} & \bf{0.71}  \\
    KG-LLaMA-13B & 0.6968  & 0.64\\
    \hline
    \end{tabular}
    \label{tab:relation_results}
    \end{table}



    \begin{table*}[t]
      \caption{Examples outputs from different LLMs. The first line is taken from FB13-100 and the second line is from YAGO3-10-100.}
    \footnotesize
    \centering
    \renewcommand{\arraystretch}{1.2}
    \begin{tabular}{c|c|p{2.4cm}|p{5.2cm}}
    \hline
    \bf{Input} & \bf{Label} & \multicolumn{2}{c}{\bf{Output}}    \\
    \hline
    \multirow{11}{*}{Is this true: Everett T Moore profession Librarian?} & \multirow{11}{*}{True} & \multirow{2}{*}{GPT-4} & I cannot verify specific personal information about individuals who are not public figures.\\\cline{3-4}
      & & \multirow{2}{*}{ChatGPT} & I'm sorry, but I don't have enough information to confirm whether Everett T Moore is a librarian or not.\\\cline{3-4}
     & & \multirow{1}{*}{LLaMA-7B} &  I was wondering if anyone could tell me if this is true. \\\cline{3-4} 
     & & \multirow{2}{*}{LLaMA-13B} &  I'm a librarian at Everett T Moore Middle School in San Antonio, TX. \\\cline{3-4} 
     & & KG-LLaMA-7B & Yes, this is true. \\\cline{3-4}
     & & KG-LLaMA-13B & Yes, this is true. \\\cline{3-4}
     & & ChatGLM-6B & Yes, Everett T Moore is a profession Librarian. \\\cline{3-4}
     & & KG-ChatGLM-6B & Yes, this is true. \\
        \hline
    \multirow{15}{*}{Josip Škorić has gender} & \multirow{15}{*}{male} & \multirow{2}{*}{GPT-4} & Josip Škorić is a male name. Josip is a Croatian form of the name Joseph, which is a masculine name.\\\cline{3-4}
      & & \multirow{3}{*}{ChatGPT} & Josip Škorić is a human being and therefore has a gender. However, I couldn't find any information about the gender of Josip Škorić.\\\cline{3-4}
     & & \multirow{2}{*}{LLaMA-7B} &  Josip Škorić has gender issues. I'm a 22-year-old guy from Croatia, who has gender issues. \\\cline{3-4} 
     & & \multirow{2}{*}{LLaMA-13B} &  Josip Škorić has undergone gender reassignment surgery. \\\cline{3-4} 
     & & KG-LLaMA-7B & male \\\cline{3-4}
     & & KG-LLaMA-13B & male \\\cline{3-4}
     & & \multirow{3}{*}{ChatGLM-6B} & As an AI language model, I do not have access to information about specific individuals' genders or their personal identity, including JosipŠkorić's.  \\\cline{3-4}
     & & KG-ChatGLM-6B & male \\
    \hline
    \end{tabular}

    \label{tab:triple_examples}
    \end{table*}

\subsection{Datasets and Settings}
We ran our experiments on four widely used benchmark KG datasets: WN11~\cite{socher2013reasoning}, FB13~\cite{socher2013reasoning}, WN18RR and YAGO3-10~\cite{dettmers2018convolutional}. Table~\ref{tab:statistics} provides statistics of all datasets we used. We used the same entity and relation text descriptions as in~\cite{yao2019kg}. Due to the access limit of GPT-4, we randomly selected 100 test examples from FB13 and YAGO3-10 for evaluation, we name the subsets FB13-100 and YAGO3-10-100.

We compare KG-LLM with multiple KG embedding methods: TransE and its extensions TransH~\cite{wang2014knowledge}, TransD~\cite{ji2015knowledge}, TransR~\cite{lin2015learning}, TransG~\cite{xiao2016transg} and TranSparse~\cite{ji2016knowledge}, DistMult and its extension DistMult-HRS~\cite{zhang2018knowledge}. The neural tensor network NTN~\cite{socher2013reasoning}. CNN models: ConvKB~\cite{SWJ318}. Contextualized KG embeddings: DOLORES~\cite{wang2018dolores}. KG embeddings with textual information: TEKE~\cite{wang2016text}, DKRL~\cite{xie2016representation} (BERT encoder), AATE~\cite{an2018accurate}. Pre-trained language models: KG-BERT~\cite{yao2019kg}, StAR~\cite{wang2021structure}, KGT5~\cite{saxena2022sequence} and KGLM~\cite{youn-tagkopoulos-2023-kglm}. We also compare with ChatGPT and GPT-4.

For instruction tuning and inference of ChatGLM-6B, We used the default parameter settings in its public implementations. For LLaMA, we use the implementation in the Transformers Python library. More detailed settings can be found in our code. For KG completion models, we use the results in their original papers or reproduce the results using default configurations in their implementations. For KGT5, we use our prompts and responses for training, other settings are the same as its implementation. We input our designed prompts to the web interface of GPT-4 and ChatGPT to obtain results.


\subsection{Results}
Table~\ref{tab:triple_results} presents triple classification accuracy scores on WN11 and FB13. If the ground truth is true and the response contains affirmative words like ``Yes" and ``yes", or if the label is false and the response contains negative words like ``No"/``no"/``not"/``n't", we label the response as correct. We find that LLaMA-7B and LLaMA-13B perform poorly on both WN11 and FB13. However, when instructed to process KG data, KG-LLaMA shows significant improvement compared to LLaMA. KG-LLaMA2-13B achieves the highest accuracy scores on the two KG data sets. Table~\ref{tab:triple_results_100} presents the accuracy scores of different LLMs on the 100 test instances of FB13. We manually label the response of different LLMs as correct or wrong. We find that KG-LLaMA performs well, the score is higher than ChatGPT and equal to GPT-4.  

The link prediction hits@1 scores of various pre-trained language models on WN18RR and YAGO3-10 are presented in Table~\ref{tab:link_results}. The scores are the average for both head and tail entities. In the case of LLMs, the response is considered correct if it contains the label words. The results indicate a promising outcome in our paradigm, as KG-LLaMA shows significant improvements due to instruction turning. Incorporating structural information also improves the results by a large margin.
Table~\ref{tab:relation_results} demonstrates that KG-LLaMA-7B produces the best relation prediction hits@1 on YAGO3-10, even better than GPT-4. KG-ChatGLM-6B also shows much better results. This indicates that instruction turning leads the LLM to extract knowledge stored in model parameters more efficiently.

Table~\ref{tab:triple_examples} illustrates the differences in responses between LLM and KG-LLM given the same input. We found the answers of the original models are not satisfactory while instruction tuning can teach the models to answer like training triples and to be more aware of a fact.

The main reasons why KG-LLM performs well are: 1). LLMs contain more general knowledge compared with smaller pre-trained language models. 2). Instruction tuning fills the gap between the pre-trained weights in LLMs and KG triple descriptions.

\section{Conclusions and Future Work}

In this work, we introduce a novel KG completion method called KG-LLM. Our approach achieves state-of-the-art performance in KG completion tasks, including triple classification and relation prediction. We aim to explore more effective prompt engineering techniques for LLMs. Additionally, we plan to enhance performance in the entity prediction task by initially selecting the top $n$ candidate entities using structural methods like TransE, followed by ranking them with LLMs.

\bibliographystyle{IEEEtran}
\bibliography{custom}

\end{document}